\documentclass[journal]{IEEEtran}

\usepackage{graphicx}
\usepackage{graphics}
\usepackage{caption}
\usepackage{lipsum}

\usepackage[T1]{fontenc}
\usepackage[utf8]{inputenc}
\usepackage[british,UKenglish,USenglish,english,american]{babel}
\usepackage{courier}
\usepackage{authblk}
\usepackage{subcaption}
\usepackage[square,numbers]{natbib}
\setcitestyle{numbers}

\usepackage{booktabs,siunitx,array,threeparttable, makecell}
\ifCLASSINFOpdf
\else
\fi
\begin{document}
%
\title{A BERT-based Deep Learning Approach for Reputation Analysis in Social Media}
%
%
%

\author[1]{Mohammad Wali Ur Rahman}
\author[1]{Sicong Shao}
\author[2]{Pratik Satam}
\author[1]{Salim Hariri}
\author[3]{Chris Padilla}
\author[3]{Zoe Taylor}
\author[3]{Carlos Nevarez}
\affil[1]{Dept. of Electrical and Computer Engr, The University of Arizona, Tucson, AZ 85721 USA}
\affil[2]{Dept. of Systems and Industrial Engr, The University of Arizona, Tucson, AZ 85721 USA}
\affil[3]{Legendary Labs, Provo, UT 84604 USA}
\maketitle

\thispagestyle{empty}
\begin{abstract}

Social media has become an essential part of the modern lifestyle, with its usage being highly prevalent. This has resulted in unprecedented amounts of data generated from users in social media, such as users’ attitudes, opinions, interests, purchases, and activities across various aspects of their lives. Therefore, in a world of social media, where its power has shifted to users, actions taken by companies and public figures are subject to constantly being under scrutiny by influential global audiences. As a result, reputation management in social media has become essential as companies and public figures need to maintain their reputation to preserve their reputational capital. However, domain experts still face the challenge of lacking appropriate solutions to automate reliable online reputation analysis. To tackle this challenge, we proposed a novel reputation analysis approach based on the popular language model BERT (Bidirectional Encoder Representations from Transformers). The proposed approach was evaluated on the reputational polarity task using RepLab 2013 dataset. Compared to previous works, we achieved 5.8\% improvement in accuracy, 26.9\% improvement in balanced accuracy, and 21.8\% improvement in terms of F-score.

\end{abstract}


%
\IEEEpeerreviewmaketitle

\section{Introduction}
%
%
%
%

In recent years, Online Reputation Management (ORM) has become an integral part of marketing strategies for influential entities, such as large enterprises and public figures. ORM aims to help an entity build its public image, manage its reputation, mitigate the negative impact of potential actions, and reclaim its lost reputation. Social media posts, blogs and articles, and other opinionated content about the entity significantly impact ORM analysis. One of the most famous examples of ORMs recently is Cambridge Analytica due to their data scandal issue with Facebook \cite{b40}.


To perform these duties for ORM, reputation management teams need to monitor texts, posts, and news regarding the target entity from online social media. A significant problem with traditional data sources (e.g., news articles, newspaper blogs, surveys, etc.) is that they usually present information from their own point of view, disregarding the opinions of most of the population. Besides, the traditional data source often causes a significant delay between the time an event occurs and the time news about such events is published or broadcasted \cite{b41}. Therefore, to overcome such limitations and collect public opinion regarding the entity of reputation in a direct and timely manner, ORM domain experts have focused more on online social media platforms \cite{b36, b37}.

The population's opinions are expressed in texts posted on social media platforms \cite{b50}. Thus, instantaneous reactions regarding an event can be mined from online social media posts \cite{b51, b68}. Also, the rapid growth of social media platforms has enabled users to engage in online posts regarding different entities related to the events, which anyone can see and share on the social media platform \cite{b52, b53}. In this way, the posts can circulate among a large number of social media users and form a public opinion of the entity for ORM \cite{b38}. Therefore, ORM experts place a great deal of emphasis on analyzing social media posts based on reputation polarity. Reputation Polarity for ORM can be defined as follows: Whether the content posted on the Internet influences a target entity's reputation positively, negatively, or neutrally. It is important to distinguish between sentiments and reputation polarity derived from text, as pointed out in \cite{b1} and \cite{b25} because a negative sentiment text sample might actually prove to be a positive reputation text sample for a target entity and vice versa. As an example, consider the following text, \textit{"Alas! Diego Maradona is no more! He will be missed by millions of fans."}, this text is a negative sentiment sample as it expresses sorrowfulness, but the reputation polarity of the text is positive for the entity "Diego Maradona". Thus, determining how text influences an entity's reputation is challenging.



Recent research has increasingly relied on pre-trained language models for different types of NLP tasks \cite{b22}. The advantage of pre-trained language models is that they have been trained on a large corpus of text data, which can significantly benefit NLP research. It is cheaper, faster, and easier to fine-tune such pre-trained models to perform downstream NLP tasks and achieve remarkable results with transfer learning. Hence, our approach is built upon BERT's modeling power, a popular pre-trained language model based on transformer-based deep learning \cite{b18}. Our study shows the feasibility of leveraging the contextual word representation derived from pre-trained BERT, along with a fine-tuning technique with extra reputation analysis data, to address reputation polarity task and achieve superior performance than previous works on RepLab 2013 dataset.

\indent The rest of the paper is organized as follows: Section II covers the related works, Section III discusses our BERT-based Reputation Analysis approach, Section IV presents our experimental results, and Section V concludes the paper.


\thispagestyle{empty}
\section{Related Works}
%
%
%
%
Over the past decade, there have been a few notable contributions to reputation analysis in natural language processing area. RepLab is a well-known project regarding ORM competition, and identifying the reputation polarity of tweets has been one of the major tasks of RepLab \cite{b1}. The participants in the competition were provided with a corpus of tweets collected between June 1, 2012, to December 31, 2012. The tweets were about 61 entities from four domains: automotive, banking, universities, and celebrities. The tweets were manually annotated by thirteen annotators who were constantly supervised by ORM experts. Among the 11 teams that participated in the reputation polarity task, SZTE-NLP achieved the highest Accuracy (0.685) \cite{b2}. The features used for their approach are the number of positive and negative words in the text, determined by SentiWordNet \cite{b3} containing the features such as number of negation words, number of words with character repetitions, number of capitalized words, number and the polarity of acronyms, and bigrams. The features also contained whether the entity was mentioned in the tweet and the distances between the tokens and the mentions of the entity. Cossu et al. \cite{b4} used tf-idf in association with Gini purity criteria and SVM classifier for reputation analysis. Filgueiras et al. \cite{b5} used the number of words with positive and negative sentiment polarity, the number of emoticons with positive and negative polarity, punctuation marks, and the number of capitalized words as features. Most of the participating teams used different types of sentiment lexicons to derive the sentiment polarity of the tweets with varying kinds of textual selection approaches like concept count vectors, bag of words, and terms co-occurrence representation (TCOR), KL-divergence, etc. \cite{b6, b7, b8, b9}. In addtion, Giachanou et al. \cite{b11} used a lexicon augmentation approach to compute reputation scores.

Apart from the teams participating in the RepLab competitions, some other researchers also made their contribution in the area of Reputation Analysis. Alrubaian et al. used some user profile-specific features other than features derived from sentiment analysis \cite{b10}. Using these features, the authors employed Logistic Regression and Feature Rank Naïve Bayes models to classify the texts based on reputation polarity. The authors used their own manually annotated data for the assessment. Manaman et al. \cite{b12} used a similar set of features based on sentiment and n-grams. However, these methods, such as bag-of-words, have the major disadvantage of not always capturing the semantic relation. Texts constructed with the same words can have different meanings due to the context, but the bag-of-words method will consistently treat those words identically. N-gram-based methods solve this problem to some extent, but the dimensionality of N-gram features is usually very high, resulting in the risk of model overfitting.

\begin{table*}[!ht]
\caption{The Monitored Entities from Different Domains in Our Reputation Polarity Prediction Task}
\centering

\begin{tabular}{cc|c|c|cc}
\multicolumn{2}{c|}{Automobile} & Banking         & Universities                            & \multicolumn{2}{c}{Celebrities}  \\ \hline

BMW        & Chrysler          & RBS bank        & Harvard University                      & Adele        & Coldplay          \\
Audi       & Subaru            & Barclays        & Stanford University                     & Alicia Keys  & Lady Gaga         \\
Volvo      & Ferrari           & HSBC            & The University of California, Berkeley~ & The Beatles  & Madonna           \\
Toyota     & Bentley           & Bank of America & MIT                                     & Led Zeppelin & Jennifer Lopez    \\
Volkswagen & Porsche           & Wells Fargo     & Princeton University                    & Aerosmith    & Justin Bieber     \\
Honda      & Yamaha            & PNC             & Columbia University                     & Bon Jovi     & Shakira           \\
Nissan     & KIA               & Capital One     & Yale University                         & U2           & PSY               \\
Fiat       & Ford              & Banco Santander & The Johns Hopkins University            & AC/DC        & The Script        \\
Suzuki     & Jaguar            & Bankia          & New York University                     & The Wanted   & Whitney Houston   \\
Mazda      & Lexus             & BBVA            & University of Oxford                    & Maroon 5     & Britney Spears    \\
           &                   & Goldman Sachs   &                                         &              &

\end{tabular}
\end{table*}

\section{BERT-based Reputation Analysis Approach}
%
%
%
%

In this section, we first review the RepLab 2013 dataset that focuses on monitoring the reputation of entities such as companies, organizations, and celebrities on Twitter. After that, we describe our BERT-based reputation analysis approach, including text preprocessing, pre-trained BERT model, and BERT-based reputation polarity.


\begin{table}[htbp]
\centering
\caption{The Description of RepLab 2013 Dataset}
\begin{tabular}{|c|c|c|c|c|}
    \hline
        ~ & Automobile & Banking & Universities & Celebrities  \\ \hline
        No. of Entities & 20 & 11 & 10 & 20  \\ \
        Train Samples & 15,123 & 7,774 & 6,960 & 15,822  \\ 
        Test Samples & 31,785 & 16,621 & 14,944 & 33,498  \\ 
        Total Samples & 46,908 & 24,395 & 21,904 & 49,320  \\  
        English Tweets& 38,614 & 16,305 & 20,342 & 38,283  \\  
        Spanish Tweets & 8,294 & 8,090 & 1,562 & 11,037  \\  \hline
    \end{tabular}
\end{table}

\noindent

\thispagestyle{empty}
\subsection{Dataset}



ORM monitoring involves constant scrutiny of online (especially social) media for information relating to the entity. The focus is on the opinions, news, and events of a given entity. It aims to detect any potential threat to its reputation early, that is, opinions and issues that could harm the entity’s public image. This implies frequent examination of the latest online information. Social networks, especially Twitter, provide a key source of information in ORM monitoring. Therefore, RepLab 2013 dataset is selected for our research due to the fact that it focuses on monitoring company and individual reputations.

The RepLab 2013 dataset \cite{b1} contains tweets regarding 61 entities from four domains that are automotive, banking, universities, and celebrities. Table I provides the list of all the monitored Entities from the RepLab 2013 dataset, addressing the diversity of the dataset. Table II summarizes the description of the corpus, as well as the number of tweets for both training and test sets, and the distribution by language. Also, the distributions of training and testing samples across the four domains are shown in Table II. As we can see, the majority of the tweets are in the English language, while there are some tweets in Spanish as well. 

The domains for reputation analysis were carefully selected for the following reasons. The automotive domain was selected for analyzing the entities where the reputation largely depends on the products. The banking domain was chosen as the representative of the entities where the transparency and ethical side of the activity play a huge role in determining their reputation. The university domain was considered due to these entities whose reputation largely depends on intangible products. Celebrities' domain was targeted because their reputation greatly depends on their character and services.

\begin{figure}[htp]
    \centering
    \captionsetup{justification=centering}
    \includegraphics[width=8cm]{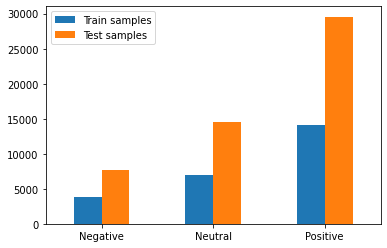}
    \caption{The reputation polarity class distribution in the dataset.}
    \label{fig:bar}
\end{figure}

Each tweet is manually annotated in three polarity classes: POSITIVE reputation, NEGATIVE reputation, and NEUTRAL reputation. In this research, all of the tweets marked as "Unrelated" are ignored as per RepLab 2013 instructions. Some of the tweets in the dataset are not available because the user have deleted the tweet or changed the Twitter account privacy settings. The distribution of resulting tweets from each polarity class is presented by Figure 1.

%
%
%
%

\subsection{Pre-trained BERT Model}

A key innovation in contextualized representation learning is BERT \cite{b18}. This work shows that even though the word embedding layer (of a traditional deep learning model for NLP task) is trained from large language corpora, training a variety of neural network architectures that encode contextual representations only based on the limited supervised data on end NLP tasks is still not sufficient. As opposed to ELMo \cite{b16} and ULMFiT \cite{b69}, which aim to add more features for a particular architecture based on human understanding of the end NLP task, BERT uses a fine-tuning process that almost does not require a specific architecture for each end NLP task. This is a desirable outcome for automated reputation analysis, as an artificial intelligence-based reputation management system aims to minimize the use of prior human knowledge in the design of the model. Instead, such knowledge should be learned through data.

BERT provides two parameter-intensive settings that are BERT-Base and BERT-Large. BERT-Base has 768 hidden dimensions, 12-layer transformer blocks, and 12 attention heads, with 110 million parameters, while BERT-Large has 1024 hidden dimensions, 24-layer transformer blocks, and 16 attention heads, with 340 million parameters \cite{b18}. Pre-training with BERT models involves two major methods: masked language modeling and next sentence prediction. The masked language modeling method randomly masks 15\% of tokens in the input, and the goal is to predict the masked token according to its context. In the next sentence prediction, pairs of sentences are used as input, and BERT predicts if the second sentence in the pair is the following sentence. Using these two methods, BERT can build highly effective bidirectional contextual representations in order to advance the performance for a variety of NLP tasks. Hence, due to the success of BERT models in NLP, we propose a BERT-based reputation polarity approach.



\begin{figure*}[htp]
    \centering
    \captionsetup{justification=centering}
    \includegraphics[width=17cm]{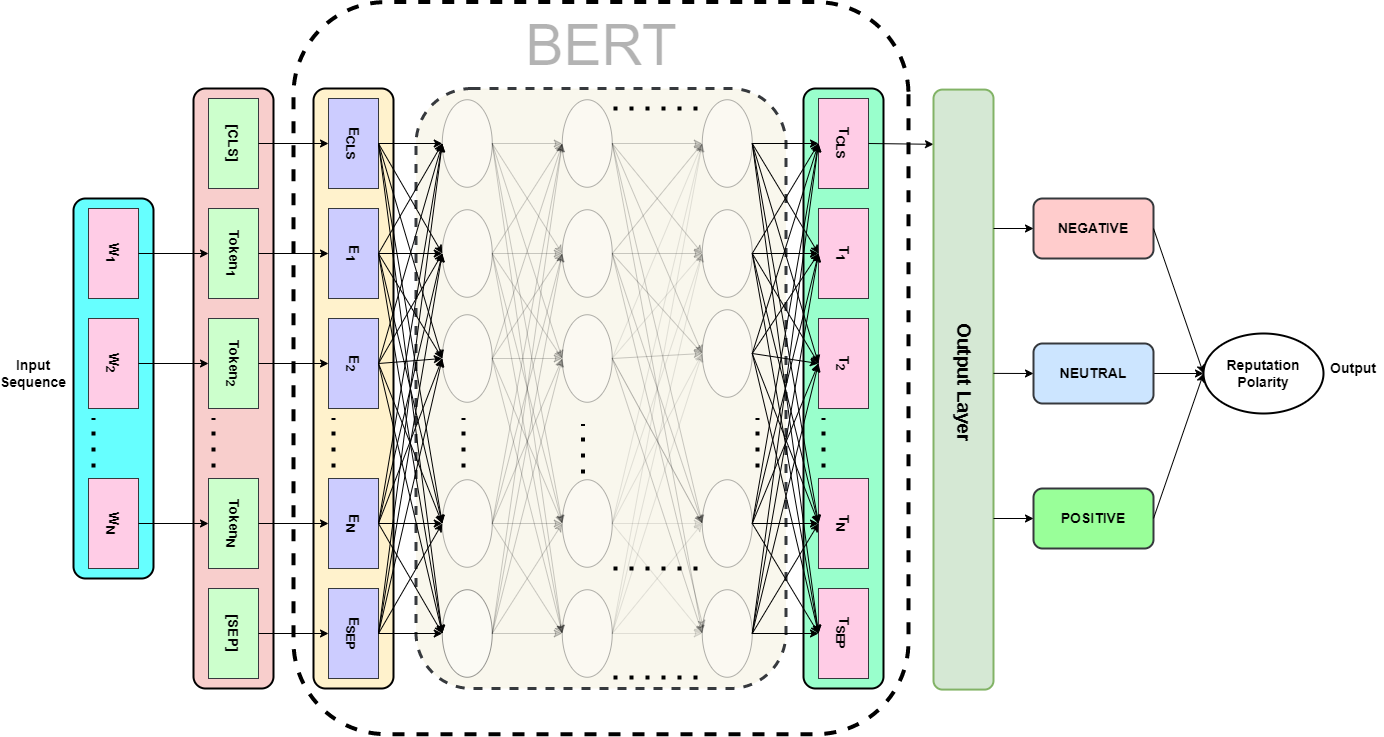}
    \caption{BERT architecture for reputation polarity task.}
    \label{fig:galaxy}
\end{figure*}


\thispagestyle{empty}
\subsection{Text Pre-processing}

Text preprocessing is an essential step for reputation polarity tasks. It transforms the text in social media into a more digestible format that is more conducive to machine learning models. With this process, tweets from RepLab 2013 dataset are cleaned and prepared for training models. Specifically, we remove extra spaces, special characters, emojis, URL links, and punctuation marks. 



\subsection{BERT-based Reputation Polarity}

We address text reputation polarity prediction in social media, which is the core task of ORM. Reputation polarity determines whether a text's content has positive, negative, or neutral implications for the monitored entity’s reputation. The traditional approaches to this task include using sentiment lexicons and linguistic features \cite{b1}. However, they struggled in social media due to its noisy nature and, most importantly, the fact that reputation polarity is not only expressed in sentiment-bearing words but also impacted by other word usage related to reputation and context. Therefore, the BERT-based approach that can learn discriminative contextual word representation, has excellent potential for this task. Thus, we propose a BERT-based deep learning Approach for building the reputation polarity classifier. We address predicting reputation polarity as a supervised task due to the label being available in the training data. PyTorch \cite{b21} platform and Transformers library from HuggingFace \cite{b54} are used for the implementation of our approach.

Figure 2 describes BERT-based reputation polarity architecture. As we can see, the BERT uses the token representation vectors $\emph E_i$ as input. For each token, $\emph E_i$ is the sum of three representation vectors, including a position embedding vector, a sentence vector, and a typical word embedding vector. The position embeddings provide the model with information about the token's location within the sentence since transformers lack this information. The sentence vector is used when more context is needed than a single sentence. A typical word embedding provides vector representations for words in the context. 
Each input token $\emph E_i$ is represented by representation $\emph T_i$ in the layers of transformer blocks by considering all token representations within the sentence, addressing capability for capturing long-term dependencies of transformer-based models. 

The BERT augments input sentence with adding [CLS] token and [SEP] token. The [CLS] token contains the embedding for special classification, and the [SEP] token is utilized for separating segments. For our reputation polarity task, BERT uses the last hidden state \textit{h} from the first token [CLS] to represent the whole input sequence. The softmax classifier is added on the top of BERT to predict the probability of reputation polarity class $c$ :
\begin{equation}
p(c|h) = softmax(Vh) 
\end{equation}where \textit{V} is the parameter matrix for specific reputation polarity prediction task. We perform fine-tuning on all the parameters of BERT and parameter matrix \textit{V} jointly learned during the fine-tuning on the reputation polarity training data.

\begin{figure*}[!htb]
     \begin{subfigure}[b]{0.5\textwidth}
         \centering
         \includegraphics[width=6 cm]{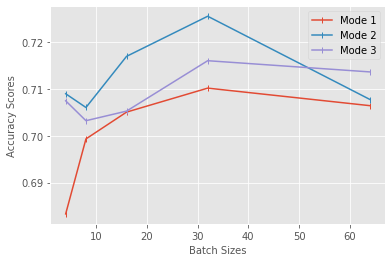}
         \caption{Accuracy Scores of different modes across different batch sizes}
         \label{fig:y equals x}
     \end{subfigure}
     \hfill
     \begin{subfigure}[b]{0.5\textwidth}
         \centering
         \includegraphics[width=6 cm]{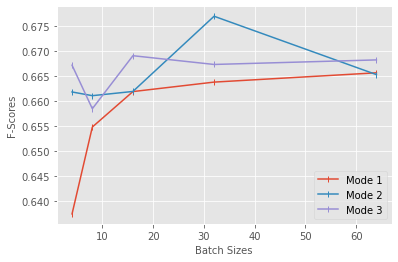}
         \caption{F-Scores of different modes across different batch sizes}
         \label{fig:three sin x}
     \end{subfigure}
     \begin{subfigure}[b]{0.5\textwidth}
         \centering
         \includegraphics[width=6 cm]{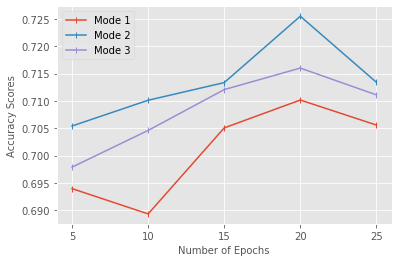}
         \caption{Accuracy Scores of different modes across different epochs}
         \label{fig:y equals x}
     \end{subfigure}
     \hfill
     \begin{subfigure}[b]{0.5\textwidth}
         \centering
         \includegraphics[width=6 cm]{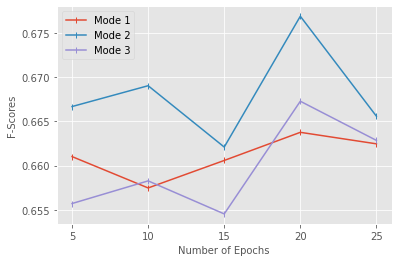}
         \caption{F-Scores of different modes across different epochs}
         \label{fig:three sin x}
     \end{subfigure}
     \begin{subfigure}[b]{0.5\textwidth}
         \centering
         \includegraphics[width=6 cm]{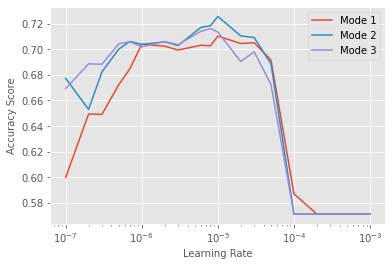}
         \caption{Accuracy Scores of different modes across different learning rates}
         \label{fig:y equals x}
     \end{subfigure}
     \hfill
     \begin{subfigure}[b]{0.5\textwidth}
         \centering
         \includegraphics[width=6 cm]{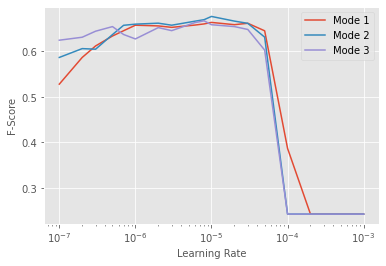}
         \caption{F-Scores of different modes across different learning rates}
         \label{fig:three sin x}
     \end{subfigure}
        \caption{Performance of our approach with different modes.}
        \label{fig:three graphs}
\end{figure*}

\begin{table*}[!htb]
    \centering
    \caption{The Best Experiment Results of Our BERT-based Reputation Analysis Approach with Different Modes}
    \begin{tabular}{ccccccccccccc}
    \toprule
        Model & Epochs & Learning\_rate & Batch\_size & Precision & Recall & F-Score & Accuracy & \thead{Training time \\ in minutes} & \thead{Test time \\ in minutes}  \\ \hline
        \midrule
        BERT (Mode 1) & 20 & $1\cdot{10^{-5}}$ & 32 & 0.68 & 0.65 & 0.66 & 0.71 & 97.1 & 3.2  \\ 
        BERT (Mode 2) & 20 & $1\cdot{10^{-5}}$ & 32 & 0.69 & 0.67 & 0.67 & 0.73 & 147.9 & 3.5  \\
        BERT (Mode 3) & 20 & $8\cdot{10^{-6}}$ & 32 & 0.69 & 0.65 & 0.67 & 0.72 & 108.9 & 3.6\\ \hline
        \bottomrule
    \end{tabular}
\end{table*}

As we can see from Figure 1, class-imbalance data occurs in our reputation polarity prediction task. The data of the minority class is significantly less than the data of the majority class. The prediction performance may be suboptimal without addressing class imbalance. To deal with this problem, a cost-based strategy \cite{b55} is employed because it can be seamlessly integrated into our BERT-based reputation polarity model. More specifically, the weight is attributed to each training sample during the training, and larger weights are applied to the class with smaller amounts, which are the training samples with higher cost if they are classified incorrectly. In contrast, lower weights are assigned to the class with larger amounts, which are the training samples with lower costs if they are wrongly classified.

\section{Experimentation Results}

\subsection{Experiment Settings}

\thispagestyle{empty}

To demonstrate the superiority of our approach, we evaluated it through RepLab 2013 dataset. As the dataset is multilingual and contains tweets from both English and Spanish Languages, we conduct the experiments by using three different pre-trained BERT models \cite{b54}: (1) bert-base-multilingual-uncased model: BERT-Base multilingual model pretrained on the top 104 languages, (2) bert-base-uncased model: BERT-Base monolingual model pretrained on English, and (3) bert-large-uncased: BERT-Large monolingual model pretrained on English. Then three different combinations of these models are formed, namely Mode 1, Mode 2, and Mode 3.

In Mode 1, we fine-tuned the \textit{bert-base-multilingual-uncased} model (i.e., BERT Multilingual Base model) on the entire training dataset (in both English and Spanish texts) and evaluated the model's performance on the test dataset. \textit{bert-base-multilingual-uncased} tokenizer is used for text tokenization. With this setting, Mode 1 can perform reputation polarity prediction on both English and Spanish tweets with a single multilingual model.

In Mode 2, we first identified the language of the tweets using Twitter API \cite{b67}. If the language of the training tweets were English, the tweets were first tokenized with the \textit{bert-large-uncased} tokenizer. Afterwards, they were passed on to the pre-trained \textit{bert-large-uncased} model (i.e., BERT Large monolingual English model) for fine-tuning on English data to create a BERT model that can detect reputation polarity focusing on English. On the other hand, if the language of the training tweets were Spanish, they were tokenized with a \textit{bert-base-multilingual-uncased} tokenizer, and then passed on to the \textit{bert-base-multilingual-uncased} model for fine-tuning on Spanish data to build a BERT model that can predict reputation polarity focusing on Spanish. For the evaluation, the test tweets were passed to either of the models depending on their language (English or Spanish). If the language of the test tweet is English, the reputation polarity of the tweet was predicted using the fine-tuned BERT large model. On the contrary, if the identified language of the test tweet was Spanish, the multilingual BERT base model was used for reputation polarity prediction.

Mode 3 was trained in a manner similarly to Mode 2. Here we also have different BERT model for each language. However, for the English training tweets, we have used them for fine-tuning \textit{bert-base-uncased} model (i.e., BERT Base monolingual English model) instead of \textit{bert-large-uncased} model. During the evaluation, the reputaion polarity of the English test tweets were predicted using the \textit{bert-base-uncased} model, which focuses on reputation polarity in English. While, for the fine-tuning and evaluation of spanish tweets, \textit{bert-base-multilingual-uncased} model in Mode 2 was still used.
\thispagestyle{empty}

For the training procedure, we have used an Adam optimizer with weight decay from Hugging Face recommended in \cite{b23}. In addition, we have used a linear scheduler with no warm-up steps. For the other hyperparameters, the original BERT authors have made some recommendations for fine-tuning tasks \cite{b18}. Apart from the number of epochs, we have mainly used the recommended hyperparameters for our classification tasks. We perform hyperparameter search for our fine-tuning process with the following setting. The learning rates were experimented in the range from $10^{-7}$ to $10^{-3}$.  The number of epoch is selected in \{5, 10, 15, 20, 25\}, and the number of batch sizes is selected in \{4, 8, 16, 32, 64\}. For the output layer, the activation function we have used are either SoftMax or LogSoftmax. Categorical cross-entropy loss function is used during the training. The cost-based strategy is employed during the training to address the class imbalance problem. The experiments were conducted with Google Colab Pro Plus \cite{b66}.


\subsection{Results}
We report the best results for all the hyperparameter combinations and plotted the graphs in Figure 3. The evaluation metrics selected for our visualization results are F-score and Accuracy because they are the official evaluation metrics used for the RepLab 2013 competition for the Reputation polarity detection task. Table III showcases our experiment runs for the three different modes of our BERT models. From the graphs in Figure 3, we can see, that Mode 2 gives us the best results, achieving the highest accuracy of 73\% with an F-score of 67\%. Besides, our BERT model in Mode 3 achieves a 72\% accuracy and Mode 1 achieves 71\% accuracy. Figure 3 show the changes in Accuracy and F-score respectively for 3 different modes with different hyperparameters.
\thispagestyle{empty}

In Figures 3(a) and 3(b), we plotted the best results obtained with different batch sizes. We observed that batch size of 32 yield best results. In Figures 3(c) and 3(d), we plotted the best accuracy and best F-Score obtained for different epoch numbers mentioned, and for our experiments, all the mode performed best when the number of epochs was 20. Figures 3(e) and 3(f) shows the prediction performance with different learning rates, ranging from $10^{-7}$ to $10^{-3}$. As we can see, all the modes obtain better prediction with lower learning rates, ranging from $10^{-6}$ to $10^{-5}$. We also noticed that the models' performance starts to drop with higher learning rates. Besides, we observed that the accuracy of both models remains unchanged from 57\% in very high learning rates mainly because the models fail to classify text as NEGATIVE or NEUTRAL, while they classify texts as POSITIVE, which is the majority class from the training set.
From the figures, we also notice that both Mode 3 and Mode 2 attain better overall results than Mode 1. 


\begin{table}[!htbp]
    \centering
    \caption{Comparison of Different Approaches in Reputation Polarity Task}
    \begin{tabular}{|c|c|c|c|}
    \hline
        Teams & Accuracy & BAC & F-score \\ \hline
        \textbf{BERT(Mode 2)} & \textbf{0.73} & \textbf{0.66} & \textbf{0.67} \\ 
        BERT(Mode 3) & 0.72 & 0.66 & 0.67 \\
        BERT(Mode 1) & 0.71 & 0.65 & 0.66 \\
        Peetz et al. \cite{b25} & - & 0.52 & 0.55 \\ 
        SZTE-NLP \cite{b2} & 0.69 & - & 0.38 \\ 
        LIA \cite{b4}& 0.65 & - & 0.19 \\ 
        Popstar \cite{b5} & 0.64 & - & 0.37 \\ 
        UAMCLYR \cite{b8}& 0.62 & - & 0.29 \\ 
        UNED ORM \cite{b9} & 0.62 & - & 0.15 \\ 
        DIUE \cite{b7}& 0.55 & - & 0.25 \\ 
        Volvam \cite{b6} & 0.54 & - & 0.26 \\ 
        DAEDALUS \cite{b99} & 0.44 & - & 0.34 \\ \hline
    \end{tabular}
\end{table}

Table IV showcases how our appropach performs compared to other approaches that classify texts based on Reputation Polarity. All the approches listed in this table use the RepLab 2013 dataset comprised of all 61 entities for their evaluation \cite{b1,b2,b4,b5,b6,b7,b8,b9,b99,b25} under the same training and testing sets.
The best results reported for each team were presented. All of the teams participating in the RepLab 2013 competition reported their Accuracy and F-scores as per the official RepLab 2013 evaluation criteria. On the other hand, Peetz et al. \cite{b25} evaluated their systems in terms of F-score and Balanced Accuracy since the dataset is imbalanced. It is plain to see that our BERT-based approaches outperform all the other approaches in every evaluation metric. For example, in terms of Accuracy, the BERT model in Mode 2 performs better by 4\% than the best-reported Accuracy by SZTE-NLP, while the BERT models in Mode 3 and Mode 1 has 3\% and 2\% better Accuracy respectively. On the other hand, the Mode 2 and Mode 3 models surpass Peetz et al. by 14\% in terms of Balanced Accuracy and 12\% in terms of F-Score. Meanwhile, the BERT model in Mode 1 surpasses Peetz et al. by 13\% in terms of Balanced Accuracy and and 11\% in terms of F-Score. These results address that the BERT successfully learns contextual word representation for reputation analysis from a sequence, which other methods fail to achieve. Therefore, we validate the superiority of the BERT-based approach for reputation polarity.

\thispagestyle{empty}
\section{Conclusion}


Over the past few years, BERT has shown its superior performance of text representation learning in many different NLP tasks \cite{b26,b27,b28,b29,b30}. However, BERT's application to classify texts based on reputation polarity has not yet been validated. The main contribution of this paper is to shed light on BERT's capability in reputation analysis. In this research, we have fine-tuned pre-trained BERT models to classify texts based on reputation polarity prediction using a standard dataset dedicated to reputation research. The experimental results show that our BERT-based approach can achieve a 5.8\% improvement in Accuracy in detecting reputation polarity, 26.9\% improvement in Balanced Accuracy, and a 21.8\% improvement in terms of F-score. We should note that it is possible further to improve the performance with more hyper-parameter tuning. Besides, the modified versions of BERT such as RoBERTa, AlBERT, distilBERT \cite{b31,b32,b33} may further improve the prediction performance for reputation analysis, We leave these possibilities for future works.

\section*{Acknowledgment}
{This work is partly supported by National Science Foundation (NSF) research projects NSF-1624668 and NSF-1849113, (NSF) DUE-1303362 (Scholarship-for-Service), and Department of Energy/National Nuclear Security Administration under Award Number(s) DE-NA0003946.}

%





\ifCLASSOPTIONcaptionsoff
  \newpage
\fi

\end{document}